\documentclass[runningheads]{llncs}
\usepackage{graphicx}
\usepackage{multicol}  
\usepackage{multirow}

\usepackage{tikz}
\usepackage{comment}
\usepackage{amsmath,amssymb} 
\usepackage{color}

\begin{document}
\pagestyle{headings}
\mainmatter
\def\ECCVSubNumber{5291}  

\title{Efficient Adversarial Attacks for Visual Object Tracking} 


\titlerunning{Efficient Adversarial Attacks for Visual Object Tracking}
%
\author{Siyuan Liang\inst{1,2} \and
Xingxing Wei\inst{4*} \and
Siyuan Yao\inst{1,2} \and
Xiaochun Cao\inst{1,2, 3*}}
\authorrunning{S. Liang, X. Wei, S. Yao, X. Cao}
%
\institute{Institute of Information Engineering, Chinese Academy of Sciences, Beijing, China \\
\email{\{liangsiyuan, yaosiyuan, caoxiaochun\}@iie.ac.cn}\\
 \and School of Cyber Security, University of Chinese Academy of Sciences, Beijing, China
 \and Cyberspace Security Research Center, Peng Cheng Laboratory, Shenzhen 518055, China
\and Beijing Key Laboratory of Digital Media, School of Computer Science and Engineering, Beihang University, Beijing 100191, China \\
\email{\{xxwei@buaa.edu.cn\}} }
\maketitle

\begin{abstract}
Visual object tracking is an important task that requires the tracker to find the objects quickly and accurately. The existing state-of-the-art object trackers, i.e., Siamese based trackers, use DNNs to attain high accuracy. However, the robustness of visual tracking models is seldom explored. In this paper, we analyze the weakness of object trackers based on the Siamese network and then extend adversarial examples to visual object tracking. We present an end-to-end network FAN (Fast Attack Network) that uses a novel drift loss combined with the embedded feature loss to attack the Siamese network based trackers. Under a single GPU, FAN is efficient in the training speed and has a strong attack performance. The FAN can generate an adversarial example at 10ms, achieve effective targeted attack (at least 40\% drop rate on OTB) and untargeted attack (at least 70\% drop rate on OTB).
\keywords{Adversarial Attack, Visual Object Tracking, Deep Learning}
\let\thefootnote\relax\footnotetext{* Corresponding Author}
\end{abstract}

\section{Introduction}
\begin{figure}[t]
\begin{center}
\includegraphics[width=0.95\linewidth]{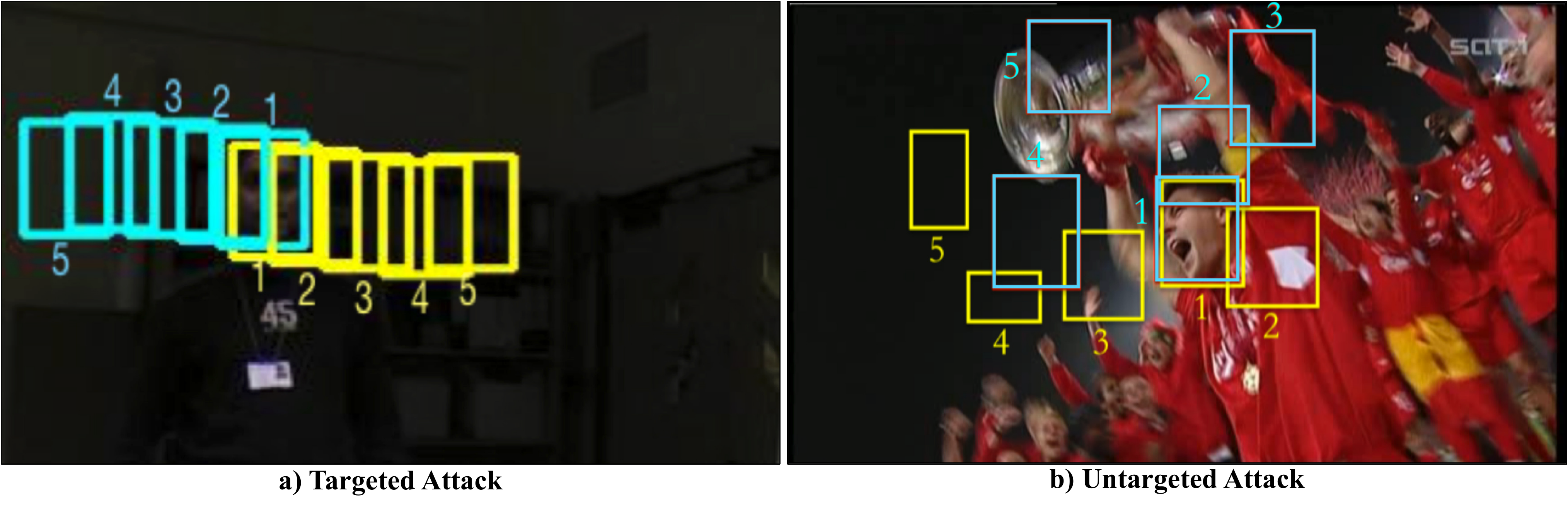}
\end{center}
   \caption{Two examples of adversarial attacks for VOT. To better show the attacking results, we plot the bounding boxes in the initial frame. The numbers represent the results in the corresponding video frames. The blue box represents the predicted bounding box, and the yellow box represents the ground truth.}
\label{fig1}
\end{figure}
Some studies have shown that DNN-based models are very sensitive to adversarial examples~\cite{goodfellow2014explaining}. In general, most recent methods for generating adversarial examples rely on the network structure and their parameters, and they utilize the gradient to generate adversarial examples by iterative optimization~\cite{szegedy2013intriguing}~\cite{carlini2017towards}. Adversarial examples have successfully attacked deep learning tasks such as image classification~\cite{moosavi2017universal}, object detection~\cite{wei2018transferable}, and semantic segmentation~\cite{fischer2017adversarial}. Researching adversarial examples can not only help people understand the principles of DNNs~\cite{wong2017provable} but also improve the robustness of networks in visual tasks~\cite{jia2019comdefend}.
\\ \indent Visual Object Tracking (VOT)~\cite{lee2015road} aims to predict the object's locations in the subsequent frames in a video when given an object’s location in the initial frame. In recent years, deep learning~\cite{danelljan2016beyond} based trackers have achieved excellent performance in many benchmarks. Among them, the SiamFC tracker~\cite{bertinetto2016fully} explores the similarity between video frames by using powerful deep features and has achieved great results in accuracy and robustness for the tracking task. Similar to the Faster R-CNN architecture~\cite{ren2015faster} in object detection, the latest visual object tracking methods are based on the Siamese network, and many variants have been derived, such as SiamVGG~\cite{yin2019siamvgg}, SiamRPN~\cite{li2018high}, SiamRPN++~\cite{li2019siamrpn++} and so on. Therefore, the significance of investigating the robustness of trackers based on deep learning becomes quite crucial.
\\ \indent In different visual tasks, the attacking targets are different. In image classification task, the target is the classification problem. In the object detection task, the target is the regression (for SSD and YOLO) or classification problem (for Faster-RCNN). In object tracking, VOT searches the most similar regions in each frame with the reference patch. Therefore the target is essentially the similarity metric problem. Thus, attacking the tracking task is totally different from the other image recognition tasks, and the existing attacking methods cannot work well (the results in Section 4.4 verify this point).
\\ \indent Regarding the above motivation, in this paper, we study the adversarial attacks on Visual Object Tracking (VOT). 
\textbf{Firstly}, because the adversarial attack on VOT is seldom explored, we give a definition of the targeted attack and untargeted attack in the visual object tracking task. 
\textbf{Then}, we propose an end-to-end fast attack network (FAN) that combines the drift loss and embedded feature loss to jointly perform the targeted and untargeted attacks on VOT. Under the hardware condition of a single GPU, we only need 3 hours off-line training on the ILSVRC15 dataset. In the inference phase, the generator can generate adversarial perturbations in milliseconds speed for the OTB dataset~\cite{wu2015object} and the VOT dataset~\cite{kristan2017visual}. Figure~\ref{fig1} gives two examples. Targeted attack causes the tracker to track object along any specified trajectory. Untargeted attack makes the tracker unable to keep track of the object.
Overall, \textbf{our contributions} can be summarized as follows:
(1)  To the best of our knowledge, we are the first one to perform the targeted attack and untargeted attack against the Visual Object Tracking (VOT) task. We analyze the weakness of the trackers based on the Siamese network, and then give a definition of the targeted attack and untargeted attack in this task.
(2) We propose a unified and end-to-end attacking method: FAN (fast attack network). We design a novel drift loss to achieve the untargeted attack effectively and apply the embedded feature loss to accomplish the targeted attack. Finally, we combine these two loss functions to jointly attack the VOT task.
(3) After three hours of training, FAN can successfully attack VOT and OTB datasets without fine-tuning network parameters. In inference, FAN can quickly produce adversarial examples within 10ms, which is much faster than iterative optimization algorithms.
\section{Related Work}
\subsection{Deep Learning in Object Tracking}
Modern tracking systems based on the deep network can be divided into two categories. The first branch is based on a tracking-by-detection framework~\cite{song2018vital}. The second branch is mainly based on SiamFC~\cite{bertinetto2016fully} and SiamRPN~\cite{li2018high}. For SiamFC, these methods focus on discriminative feature learning~\cite{zhang2019deeper}~\cite{he2018twofold}~\cite{yin2019siamvgg}, exemplar-candidate pairs modeling~\cite{dong2018triplet}, and dynamical hyperparameter optimization~\cite{dong2019dynamical}~\cite{dong2018hyperparameter}. For SiamRPN, some researchers introduce a more powerful network cascaded model~\cite{fan2019siamese} or deeper architecture~\cite{li2019siamrpn++} for region proposal searching. DaSiam~\cite{zhu2018distractor} proposes a distractor-aware training strategy to generate semantic pairs and suppress semantic distractor. In summary, the Siamese trackers show their superior performance due to the high localization accuracy and efficiency, but most of these trackers are sensitive to the adversarial pertubations of the input data. Therefore, investigating the robustness of these trackers under  adversarial attracks becomes crucial.
\subsection{Iterative and Generative Adversary}
The existing adversarial attacks are based primarily on the optimization algorithm and generation algorithm. The optimization-based adversarial attack discovers the noise's direction by calculating the DNNs' gradient within a certain limit~\cite{carlini2017towards}. I-FGSM~\cite{kurakin2016adversarial} decomposes one-step optimization into multiple small steps, and iteratively generates adversarial examples for image classification. DAG~\cite{xie2017adversarial} regards the candidate proposal for RPN~\cite{ren2015faster} as a sample, and iteratively change the proposal's label to attack object detection and segmentation. Another type of adversarial attack is based on the generator, which can quickly generate adversarial perturbations~\cite{bose2018adversarial}. GAP~\cite{poursaeed2018generative} uses the ResNet generator architecture to misclassify images of ImageNet~\cite{deng2009imagenet}. UEA~\cite{wei2018transferable} generates transferable adversarial examples by combining multi-layer feature loss and classification loss, aiming to achieve an untargeted attack in image and video detection. Due to speed limitations, adversarial attacks based on iterative optimization cannot achieve real-time attacks in the visual object tracking task.
\section{Generating Adversarial Examples}
\subsection{Problem Definition}
Let $V=\{I_{1},...,I_{i}, ...,I_{n}\}$ be a video that contains $n$ video frames. For simplicity, we take one tracking object as the example, thus $\mathcal{B}^{gt}=\{b_{1},..., b_{i},...,b_{n}\}$ is used to represent the object's ground-truth position in each frame. The visual object tracking will predict the position $\mathcal{B}^{pred}$ of this object in the subsequent frames when given its initial state. For different datasets, the predicted output is different. In general, four points $b_{i}\in\mathcal{R}^{4}$ are used to represent the box.
\\ \indent In SiamFC~\cite{bertinetto2016fully}, the tracker $f_{\theta}(\cdot)$ with parameters $\theta$ first transforms the reference frame $I_{R}$ and annotation $b^{init}$ to get an exemplar region  $z=\tau(I_{R}, b^{init})$, and searches a large area $b^{search}$ in the candidate frame $I_{C}$ to get a candidate region $x=\tau(I_{C}, b^{search})$. After feature extraction $\varphi(\cdot)$, a fully-convolutional network is used to calculate the similarity between $z$ and $x$ to get the response score map $\mathcal{S}=f_{\theta}(z, x)=\varphi(z)\ast\varphi(x)$. A Cosine Window Penalty (CWP) \cite{bertinetto2016fully} is then added to generate the final bounding box $b_{i}=CWP(\mathcal{S})$. CWP can penalize the large offset, making the predicted box not far from the previous box. 
\\ \indent $\hat{V}=\{\hat{I}_{1},...,\hat{I}_{i},...,\hat{I}_{n}\}$ represents the adversarial video. The generator mainly attacks the candidate area $\hat{x}_{i}=\tau(\hat{I}_{i}, b_{i}^{search})$ in the adversarial frame $\hat{I}_{i}$. The definitions of targeted and untargeted attacks in VOT are given below:

(1) Targeted Attack. The adversarial video $\hat{V}$ guides the tracker to track the object along the specified trajectory $\mathcal{C}^{spec}$, i.e., $\forall i, ||\hat{c}_{i}-c^{spec}_{i}||_{2}\leq\varepsilon$,s.t.~$\hat{c}_{i}=center(CWP(f(z,\hat{x}_{i})))$. $center(\cdot)$ gets the prediction center through the prediction box. The Euclidean distance between the prediction center $\hat{c}_{i}$ and the target center $c^{spec}_{i}$ should be small. Here we set $\varepsilon$ to 20 pixels.

(2) Untargeted Attack. The adversarial video $\hat{V}$ causes the adversarial trajectory $\mathcal{B}^{attack}=\{CWP(f(z,\hat{x}_{i}))\}^{n}_{i=1}$ to deviate from the original trajectory $\mathcal{B}^{gt}$ of an object. When the IOU of the prediction box and the ground-truth box is zero, i.e., $IOU(\mathcal{B}^{attack}, \mathcal{B}^{gt})=0$, we think that the untargeted attack is successful.
\subsection{Drift Loss Attack}
Trackers based on the Siamese network are highly dependent on the response map generated by the fully-convolutional network to predict the object's location. Because the SiamFC uses a search area $x$ when predicting the object's location, we can attack this search area to achieve untargeted attack. Over time, the tracker will accumulate the predicted slight offset until the tracker completely loses the object.
\begin{figure}[t]
\begin{center}
\includegraphics[width=1\linewidth]{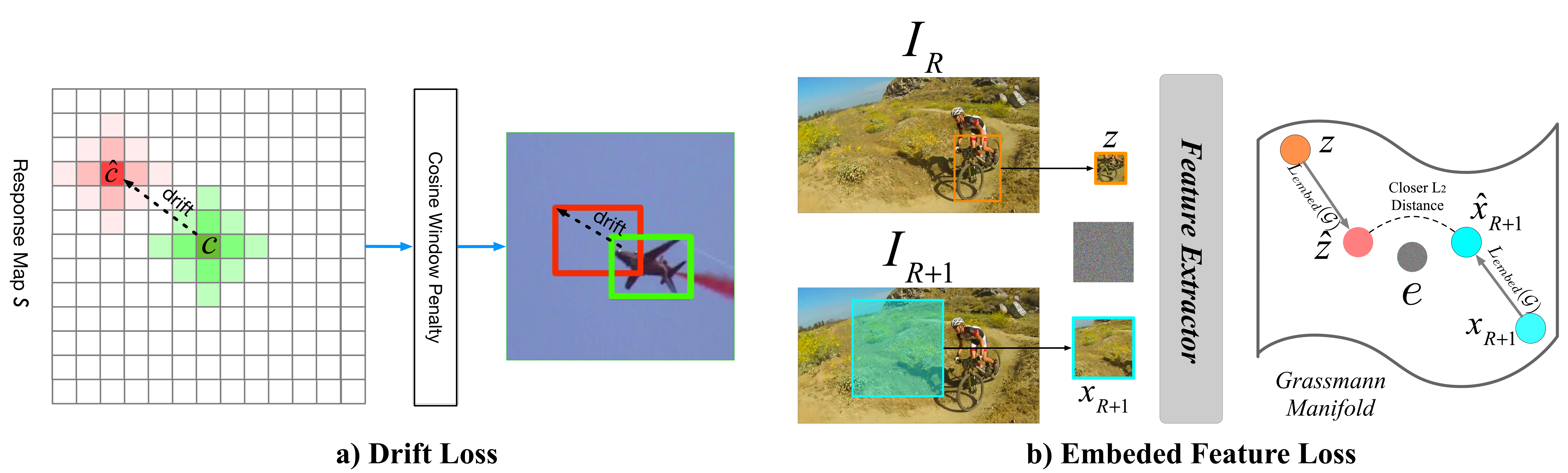}
\end{center}
   \caption{Overview of the proposed drift loss and embedded feature loss. They are designed for untargeted attack and targeted attack. For details, see sections 3.2 and 3.3.}
\label{fig3}
\end{figure}
\\ \indent In Figure \ref{fig3} a), the darker the color in response map $\mathcal{S}$, the greater the response score. The red area and green area represent the response regions of the adversarial image and clean image. $c$ represents the maximum score in the response map. For a well-trained tracker, the response map of clean examples are generally concentrated on the central area (green area). Thus, we propose a drift loss, which generates adversarial perturbations that drift the activation center of $\mathcal{S}$:
\begin{equation}
l(y,s) = \log(1+exp(-ys)),\label{eq1}
\end{equation}
where $s$ represents the response score and $y\in(-1,1)$ represents the label of grid in response map $\mathcal{S}$. The central part of the response map $\mathcal{S}$ (green area) is labeled 1, and the rest is -1. In order to generate adversarial examples, the maximum response value of the non-intermediate response map is greater than the maximum response value of the ground-truth, so the score loss of the response map can be written as:
\begin{equation}
\mathcal{L}_{score}(\mathcal{G}) = \min\limits_{p\in\mathcal{S}^{+1}}(l(y[p],s[p]))-\max\limits_{p\in\mathcal{S}^{-1}}(l(y[p],s[p])),\label{eq2}
\end{equation}
where $p\in\mathcal{S}$ represents each position in the response map. The offset of the prediction box depends on the offset of the activation center in the response map. We want the activation center to be as far away from the center as possible, so the distance loss can be expressed as:
\begin{equation}
\mathcal{L}_{dist}(\mathcal{G}) = \frac{\beta_{1}}{\delta+\|p^{+1}_{max}-p^{-1}_{max}\|_{2}}-\xi,\label{eq3}
\end{equation}
where $p_{max}^{i}=\arg\max\limits_{p\in\mathcal{S}^{i}}(s[p]), i=+1, -1$ represents position of max activation scores in positive areas or negative areas of response map. $\delta$ is a small real number, and $\beta_1$ controls weight in distance loss. $\xi$ controls the offset degree of the activation center. Usually, the activation center leaves the central area. The drift loss consisting of score loss and distance loss can be written as:
\begin{equation}
\mathcal{L}_{drift} = \mathcal{L}_{dist} + \beta_{2}\mathcal{L}_{score} .\label{eq4}
\end{equation}
\subsection{Embedded Feature Loss Attack}
Since the targeted attack requires the tracker to track along the specified trajectory, it is different from the untargeted attack. The drift loss in Section 3.2 is easy to achieve the untargeted attack, but its attack direction is random, and it cannot achieve targeted attack. The input to the targeted attack are a video $V$ and the specified trajectory's centers $\mathcal{C}^{spec}$. Due to the great difference between the object and background, the response value of the candidate image $x_{R+1}$ and the exemplar image $z$ along the specified trajectory will gradually drop to be lower in the background area. Thus, the targeted attack will soon fail. 
\\ \indent For effective targeted attack, we need increase the response value. As shown in Figure~\ref{fig3} b), we want to minimize the $L_{2}$ distance between the features of the adversarial exemplar and the specific trajectory area. Thus, we propose embedded feature loss that generates adversarial images $\hat{z}$ and $\hat{x}_{R+1}$. The features of the generated adversarial examples are close to the features of the embedded image $e$. 
\begin{equation}
\mathcal{L}_{embed}(\mathcal{G}) = \|\varphi(q+\mathcal{G}(q))-\varphi(e)\|_{2},\label{eq5}
\end{equation}
\\ \indent In Eq~\ref{eq5}, $e$ represents the specified trajectory area, $q\in\{z, x_{R+1}\}$ represents input video area. $z$ and $x_{R+1}$ represent the exemplar frame and the $R+1$ frame to track. $\varphi$ represents the feature function, and $\mathcal{G}(q)$ represents adversarial perturbation. After feature extraction, the features of the adversarial image and the embedded image should be as close as possible to achieve targeted attack.
\\ \indent In the training phase, the choice of embedded images is very important. For example, the feature distance between a shepherd dog and a sled dog is smaller than that of a shepherd dog and an Egyptian cat.  In the actual attack, we find that attacking a video frame to an object will produce significant perturbations. We use Gaussian noise to replace the object feature in $e$ to optimize Eq~\ref{eq5}, but the specified trajectory remains unchanged.
\begin{figure*}[t]
\begin{center}
\includegraphics[width=1\linewidth]{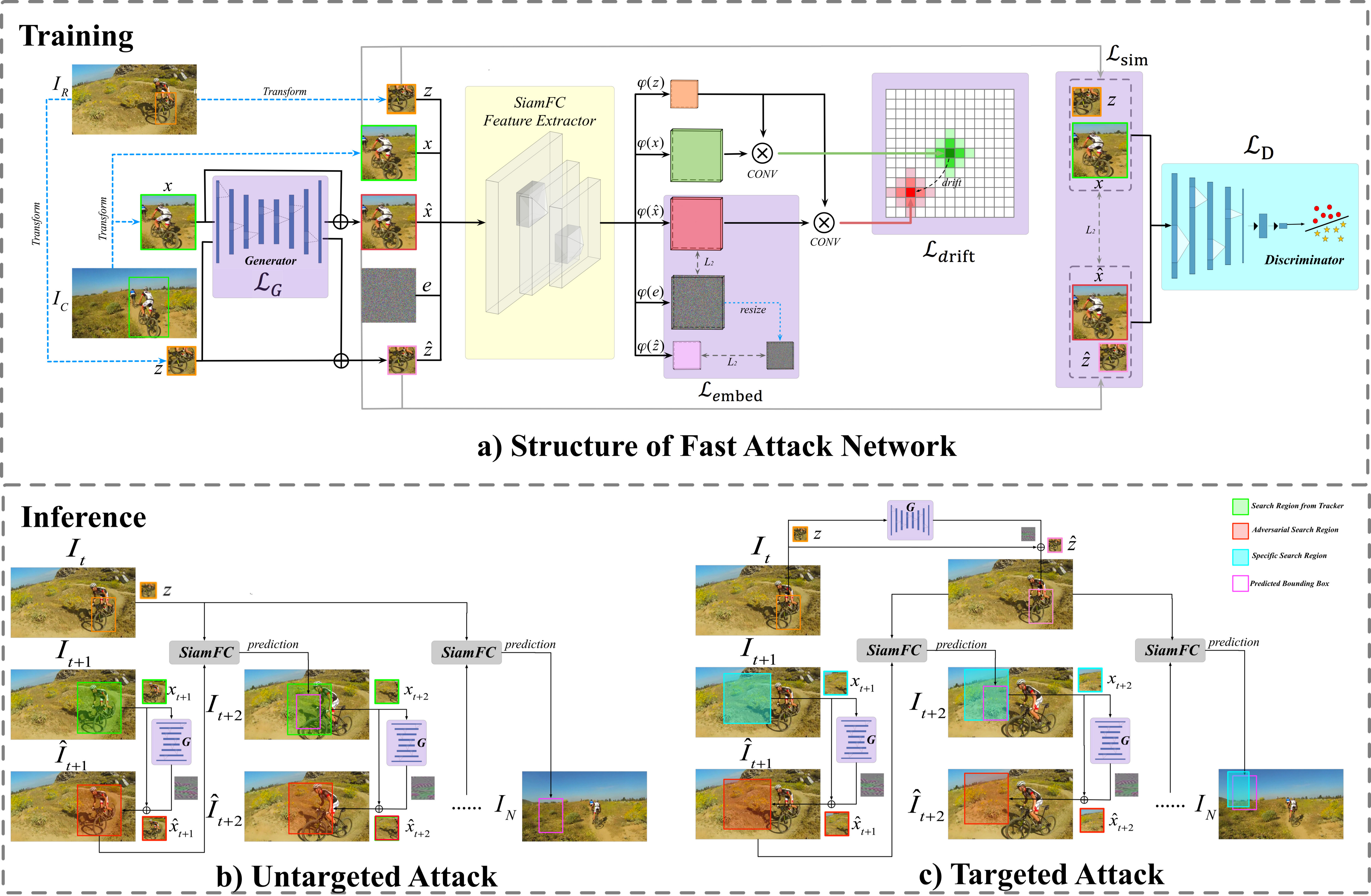}
\end{center}
 \caption{The training and inference framework of FAN. In a), we train the generator and discriminator using the well-trained SiamFC architecture (yellow area + convolution parameters). The losses of the generator and the discriminator are highlighted by the purple parts and blue parts, respectively. We can achieve both a targeted attack and an untargeted attack by adjusting the loss weight of the generator. For untargeted attack b), we only generate adversarial perturbations for search area $x$ in the candidate image $I_{C}$. For targeted attack c), we attack both the exemplar image $z$ and the specific search area (the blue part in $c$), which is determined by the specific trajectory.}
\label{fig2}
\end{figure*}

\subsection{Unified and Real-time Adversary}
As shown in Figure \ref{fig2}, we train a GAN to generate adversarial examples. Necessarily, generating adversarial perturbations can be seen as an image translation task~\cite{poursaeed2018generative}. We generate adversarial perturbations for candidate images in the candidate frames, which are more difficult to perceive in space. We refer to cycle GAN~\cite{zhu2017unpaired} as a generator to learn the mapping from natural images to adversarial perturbations. We adopt the generator proposed in paper~\cite{johnson2016perceptual} and use nine blocks to generate adversarial perturbations. For the discriminator, we use PatchGAN~\cite{isola2017image}, which uses the overlapping image patch to determine whether the image is true or false.
\\ \indent The loss of the discriminator can be expressed as:
\begin{equation}
\begin{aligned}
\mathcal{L}_{\mathcal{D}}(\mathcal{G},\mathcal{D},\mathcal{X})&=\mathbb{E}_{x \sim p_{data}(x)}[(\mathcal{D}(\mathcal{G}(x)+x))^{2}]\\&+\mathbb{E}_{x \sim p_{data}(x)}[(\mathcal{D}(x)-1)^{2}].
\end{aligned}
\label{eq6}
\end{equation}

In the training phase, we  train the discriminator by minimizing Eq~\ref{eq6}. In order to make the image generated by the generator more realistic, the loss of the generator can be expressed as:
\begin{equation}
\begin{aligned}
\mathcal{L}_{\mathcal{G}}(\mathcal{G},\mathcal{D},\mathcal{X})=\mathbb{E}_{x \sim p_{data}(x)}[(\mathcal{D}(\mathcal{G}(x)+x)-1)^{2}].
\end{aligned}
\label{eq7}
\end{equation}

In addition, we use the $L_{2}$ distance as a measure to minimize the loss of similarity so that the adversarial image is closer to the clean image in visual space. The loss of similarity can be expressed as:
\begin{equation}
\mathcal{L}_{sim}(\mathcal{G}) = \mathbb{E}[\| \mathcal{X}-\mathcal{\hat{X}} \|_{2}].\label{eq8}
\end{equation}

Finally, the full objective for the generator can be expressed as:
\begin{equation}
\begin{aligned}
\mathcal{L} = \mathcal{L}_{\mathcal{G}} + \alpha_{1}\mathcal{L}_{sim}+\alpha_{2}\mathcal{L}_{embed}+\alpha_{3}\mathcal{L}_{drift},
\end{aligned}
\label{eq9}
\end{equation}
We propose a \textbf{unified network} architecture, which can achieve a targeted attack and untargeted attack by adjusting the hyperparameters. $\beta_{1}, \beta_{2}$ make $L_{dist}$ and $L_{score}$ roughly equal. Thus, there is no need for special adjustment. $\xi$ controls the offset degree of the activation center. $\alpha_{1}$ and $\alpha_{3}$ control the untargeted attack. We fix $\alpha_{3}$ and adjust $\alpha_{1}$ from the visual quality. $\alpha_{2}$ controls embedding image features. We test value from 0.05-0.1 and the precision score improves ten percentage. For the targeted attack, we do not need drift loss, so set $\alpha_{3}$ to 0, $\alpha_{1}=0.0024$ and $\alpha_{2}=0.1$. For the untargeted attack, we set $\alpha_{2}$ to 0, $\alpha_{1}=0.0016$, and $\alpha_{3}=10$. In Eq~\ref{eq3}, we set $\beta_{1}=1$, $\delta=1*10^{-10}$, $\xi=0.7$. In Eq~\ref{eq4}, $\beta_{2}$ is set to 10. We use Adam algorithm~\cite{mathieu2015deep} to optimize generator $\mathcal{G}$ and discriminator $\mathcal{D}$ alternatively. Using a GPU Titan XP, we can get the best weight by iterating about 10 epochs(about 3 hours) on the ILSVRC 2015 dataset. 
\\ \indent Since the prediction box of the tracker in the current frame is strongly dependent on the results of the previous frame, we can make the prediction box produce a small error offset and eventually stay away from the ground-truth trajectory. We only add perturbations to the candidate image $x$ for the untargeted attack. For targeted attack, we embed features of embedding images in exemplar states $z$ and candidate images $x$ by adding adversarial perturbations. Although the adversarial attack deals with a large number of videos, the generator can generate adversarial examples in milliseconds. This enables us to complete the \textbf{real-time adversarial attack} for visual object tracking.

\section{Experiments}
\subsection{Datasets and Threat models}
We train the generator and discriminator on the training set of the ILSVRC 2015 dataset~\cite{russakovsky2015imagenet}. We refer to the training strategy in SiamFC~\cite{bertinetto2016fully}. After training is completed, the generator is tested on four challenging visual object tracking datasets without parameter adjustment: OTB2013~\cite{wu2013online}, OTB2015, VOT2014, and VOT2018~\cite{kristan2017visual}. Specifically, the VOT datasets will be re-initialized after the tracker fails to track. Therefore, it is more difficult to attack VOT datasets than OTB datasets. We use SiamFC based on Alexnet as a white-box attack model. SiamRPN~\cite{li2018high}, SiamRPN+CIR~\cite{zhang2019deeper} and SiamRPN++~\cite{li2019siamrpn++} as black-box attack models. 

\subsection{Evaluation Metrics}
Since the targeted attack and untargeted attack are different, we define their own evaluation criteria, respectively.
\\ \textbf{Untargeted Attack Evaluation}: In the OTB dataset, we use success score, precision score, and success rate as the evaluation criteria. The \textbf{success score} calculates the average IOU of the prediction box and the ground-truth. The \textbf{precision score} indicates the percentages of the video frames whose euclidean distance between the estimated centers and ground-truth centers is less than the given threshold. The percentage of successful attacked frames to all the frames is the \textbf{success rate}.
\\ \indent In the VOT dataset, we measure the accuracy in the videos using the \textbf{success score}. Considering the restart mechanism in the VOT dataset, robustness is a more important evaluation metric. \textbf{Mean-Failures} refer to calculating the average number of failures for the object tracking algorithm in all datasets. 
\\ \textbf{Targeted Attack Evaluation}: The target attack requires the tracker to move according to a specific trajectory, so we use the \textbf{precision score} as the evaluation criteria. The higher the precision score, the more effective the targeted attack.
\\ \textbf{Image Quality Assessment}: We use \textbf{Mean-SSIM} to evaluate the quality of adversarial videos. Mean-SSIM calculates the average SSIM of frames in videos. The generated adversarial perturbations are difficult to be found when Mean-SSIM is close to 1.

\subsection{Untargeted Attack Results}
\begin{table}
\begin{center}
\begin{tabular}{|c|c|c|c|c|}
\hline
\multicolumn{2}{|c|}{Datasets} &Clean Videos & Adversarial Videos & Drop Rate  \\ \hline
\multirow{4}{*}{OTB2013}   & Success Score     & 0.53                          & 0.14                 & 74\%                       \\ \cline{2-5} 
                           & Precision Score   & 0.71                          & 0.17                 & 76\%                       \\ \cline{2-5} 
                           & Success Rate      & 0.66                          & 0.12                 & \textbf{81\%}                       \\ \cline{2-5} 
                           & Mean-SSIM              & 1                             & 0.93                 & 7\%                        \\ \hline
\multirow{4}{*}{OTB2015}   & Success Score     & 0.53                          & 0.15                 & 72\%                       \\ \cline{2-5} 
                           & Precision Score   & 0.72                          & 0.18                 & 75\%                       \\ \cline{2-5} 
                           & Success Rate      & 0.66                          & 0.12                 & \textbf{81\%}                       \\ \cline{2-5} 
                           & Mean-SSIM              & 1                             & 0.93                 & 7\%                        \\ \hline
\multirow{3}{*}{VOT2014}   & Success Score     & 0.54                          & 0.42                 & 22\%                       \\ \cline{2-5} 
                           & Mean-Failures        & 28                            & 112                  & 300\%                      \\ \cline{2-5} 
                           & Mean-SSIM              & 1                             & 0.94                 & 6\%                        \\ \hline
\multirow{3}{*}{VOT2018}   & Success Score     & 0.49                          & 0.42                 & 14\%                       \\ \cline{2-5} 
                           & Mean-Failures        & 48                            & 246                  &\textbf{413\%}                      \\ \cline{2-5} 
                           & Mean-SSIM              & 1                             & 0.97                 & \textbf{3\%}                        \\ \hline
\end{tabular}
\end{center}
\caption{Untargeted attacks on VOT and OTB datasets. We use drop rate to measure the attack performance. Large Mean-Failures means the tracker frequently lost objects.}
\label{tab1}
\end{table} 
In Table~\ref{tab1}, we report the results of the untargeted attack on four tracking datasets. The second and the third columns represent the object tracking results of SiamFC on the clean video and the adversarial video. The drop rate of tracking evaluation metrics for OTB datasets has fallen by at least $72\%$, indicating that our attack method is effective. For the quality assessment, the highest drop rate is only $7\%$, which is sufficient to show that adversarial perturbations generated by our attack method are visually imperceptible.
\\ \indent We find that the success rate is the most vulnerable evaluation metrics on the OTB dataset, with a drop rate of $81\%$. This indicates that our attack method can effectively reduce the IOU between the prediction box and the ground-truth box. Our attack method increases the number of tracking failures as high as $413\%$ on the VOT2018 dataset. Therefore, our attack method can still effectively fool the tracker and cause it to lose objects under the reinitialization mechanism. However, compared with the OTB datasets, there is no significant decrease in the success score on VOT datasets. The reason may be that the success score is still high because the tracker keeps reinitializing the object. According to the definition of untargeted attack, Mean-Failures is more reasonable for evaluating adversarial attacks. Finally, the VOT2018's Mean-SSIM dropped only $3\%$. Our generator sparsely attacks video frames over time, resulting in that perturbations less difficult to be perceived.
\begin{figure}[t]
\begin{center}
\includegraphics[width=1\linewidth]{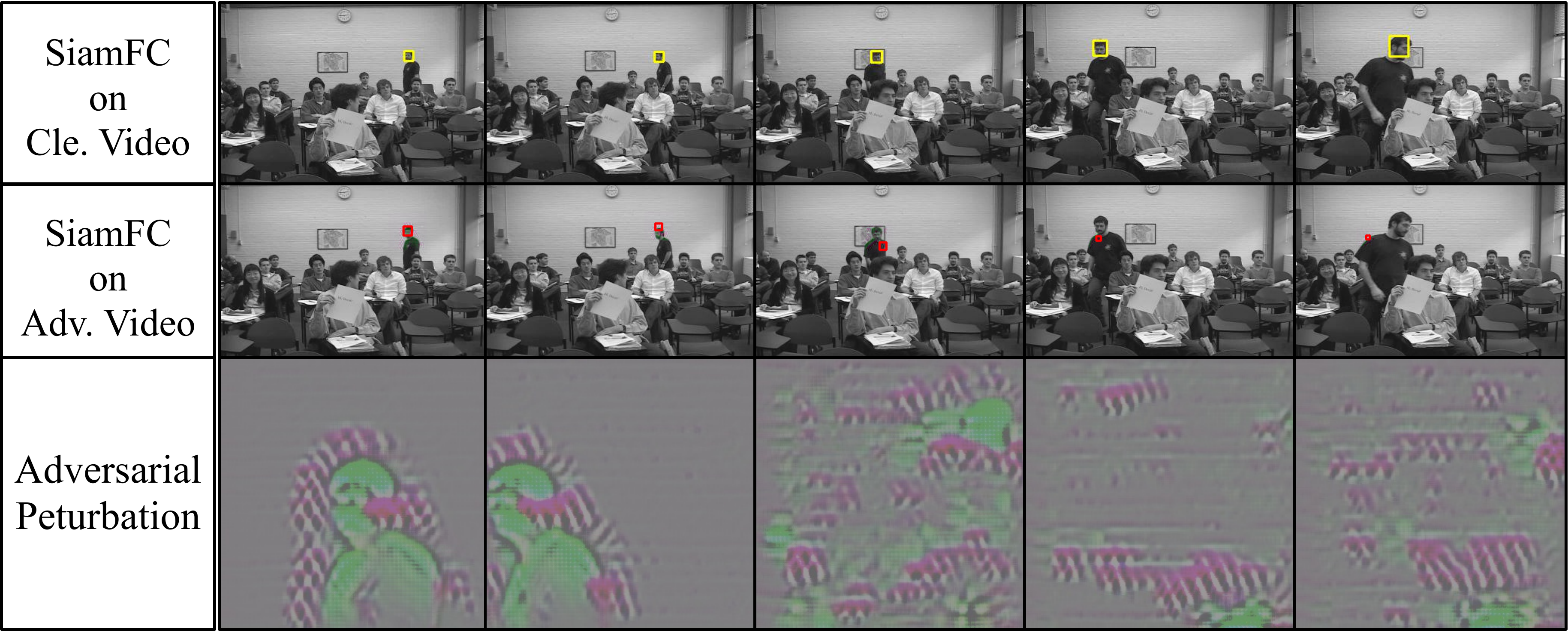}
\end{center}
\caption{We visualize the tracking results under the untargeted attack. Yellow represents a ground-truth bounding box and, red represents predicted bounding box by trackers.}
\label{fig4}
\end{figure} 
\begin{figure}[t]
\begin{center}
\includegraphics[width=0.95\linewidth]{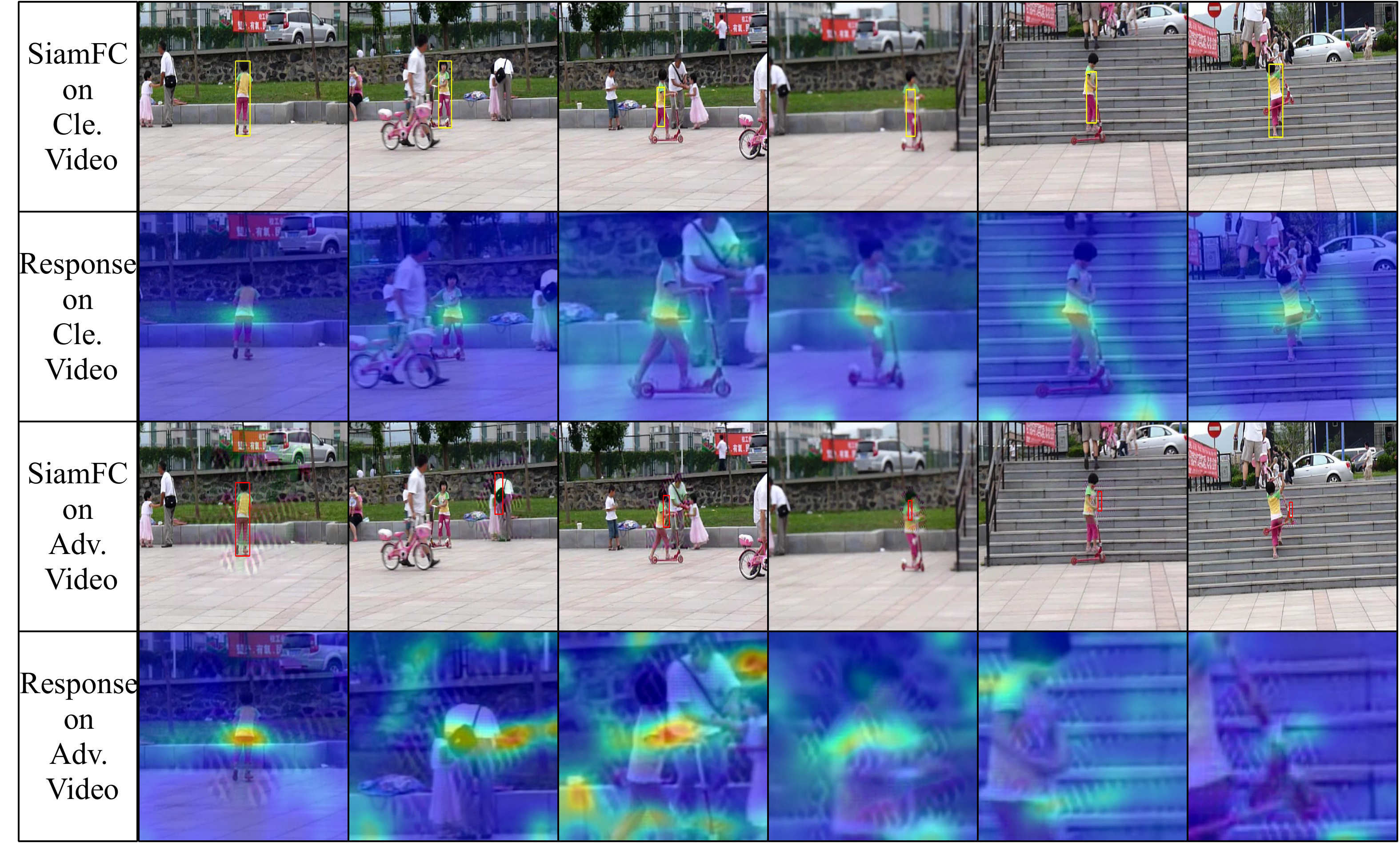}
\end{center}
\caption{The visualization of response maps between adversarial examples and clean videos, respectively. Blue indicates low response, and red indicates high response.}
\label{fig5}
\end{figure}
\\ \indent We show an adversarial video in Figure~\ref{fig4}, which is sampled equidistantly in time from left to right. We added slight perturbations in search images to successfully fool the SiamFC tracker. This kind of attack method does not produce too much deviation in a short-time and is difficult to be detected by trackers. The third line represents adversarial perturbations, and FAN can adaptively attack the critical feature areas without prior shapes. 
\\ \indent The left-to-right in Figure~\ref{fig5} are the results of the uniform sampling of a video over time. By comparing the second row and the fourth row, we can see that the responding area of the clean image is concentrated, and the scores are not much different (the green part). However, the adversarial examples generated by FAN start to cause a large range of high scores in the response map and are relatively scattered. These scattered high-scoring areas will fool the SiamFC tracker to make it impossible to distinguish the object. Due to incorrect activation of the response map, the search areas in adversarial examples will gradually shrink over time. The subsequent adversarial perturbations will also increase the degree of narrowing of the search areas (the extent of the fourth line is reduced differently in equal time). The perturbations gradually decrease in space over time due to the FAN attack on the search areas.
\subsection{Comparisons with the Baselines}
To better show the performance, we compare our FAN method with the widely used FGSM \cite{goodfellow2014explaining} and PGD \cite{madry2017towards}. The results are shown in Table 2.
\begin{table}[]
\begin{center}
\begin{tabular}{|c|c|c|c|c|c|}
\hline
Methods & Success Score & Precision Score & Success Rate & Mean-SSIM & Time(s) \\ \hline
FGSM & 3\% & 2\% & 3\% & 0.95 & 0.03 \\ \hline
PGD & 3\% & 2\% & 3\% & 0.97 & 3.53 \\ \hline
FAN & \textbf{74\%} & \textbf{76\%} & \textbf{81\%} & \textbf{0.94} & \textbf{0.01} \\ \hline
\end{tabular}
\end{center}
\label{tab41}
\caption{The untargeted attacks on OTB2013. Compared with the FAN method, the modified FGSM and PGD methods cannot achieve effective attacks. The percentage represents the drop rate compared to clean video.}
\end{table}
\\ \indent Because FGSM and PGD are used to attack the image classification task, and cannot directly attack the visual object tracking task. Therefore we make some modifications. In object tracking, tracker searches the most similar regions in each frame with the reference patch. The most similar regions in the response map are labeled 1; the others in the response map are -1. Therefore, for FGSM and PGD, the attack target is to change the correct label in the response map (invert label 1 to -1). We perform experiments on the modified FGSM and PGD methods at OTB2013 and compared them with the FAN method.
\\ \indent The percentages in Table 2 represent the drop rate versus different metrics. We can see that these two methods are not effective for attacking VOT tasks. Besides, the average time for PGD to process a sample is 3.5s, which is not suitable for attacking a large number of frames. Under the same hardware conditions, our method process a sample only need 0.01s, and it can effectively attack clean videos.
\subsection{Targeted Attack Results}
We need to set specific trajectories for the video frames in the dataset to achieve targeted attack. Since the VOT datasets will be reinitialized when the tracker is far away from the ground-truth, there is no point in implementing a targeted attack on the VOT datasets. Our targeted attack method still works because it can cause the tracker to restart multiple times on the VOT dataset. For clean videos in the VOT2014, SiamFC will restart tracking per 108.8 frames. After attacked by our method, SiamFC will restart tracking per 14 frames, which shows our method significantly increases numbers of restart for tracker in the VOT dataset in the targeted attacks.
\begin{table}
\begin{center}
\begin{tabular}{|c|c|c|c|c|}
\hline
\multicolumn{2}{|c|}{Datasets} & Clean Videos & Adversarial Videos &Drop Rate\\ \hline
\multirow{2}{*}{OTB2013}   & Precision Score   & 0.69                & 0.41                &40.6\% \\ \cline{2-5} 
                           & Mean-SSIM         & 1                   & 0.92                &8\% \\ \hline
\multirow{2}{*}{OTB2015}   & Precision Score   & 0.71                & 0.42                &40.8\% \\ \cline{2-5} 
                           & Mean-SSIM         & 1                   & 0.92                &8\% \\ \hline
\end{tabular}
\end{center}
\caption{An overview of the targeted attack results. We use precision scores to evaluate targeted attacks. A high precision score means that the tracker's prediction is close to the specified trajectory.}
\label{tab2}
\end{table}
\\ \indent We conduct experiments on OTB2013 and OTB2015 datasets. Manually labeling specific trajectories on these datasets will be time-consuming. Therefore, we generate specific trajectories based on the original annotations. Here we consider the most difficult case of a targeted attack. That is, the generated specific trajectory is completely opposite to the original trajectory. We use the following rules to calculate the bounding box for specific trajectory:
\begin{equation}
b^{spec}_{t}=
\begin{cases}
b^{gt}_{0}& \text{$t=1$}\\
2*b^{gt}_{t-1}-b^{gt}_{t}& \text{$t \geq 2$},
\end{cases}
\label{eq10}
\end{equation}
where $b^{spec}$ represents the bounding box of specified trajectory, and $b^{gt}$ represents ground-truth in datasets. 
\\ \indent In Table~\ref{tab2}, the first and second columns represent precision scores of tracker's predicted trajectory on clean videos and adversarial videos. Experiment results show that the tracking system after the targeted attack cannot reach the same precision scores on the clean video. The reason for this result may be that the automatically generated specific trajectory is not the best path that the targeted attack can choose. Even if the targeted attack of visual tracking is more difficult than an untargeted attack, FAN can still successfully attack most videos under the most difficult specific trajectories.
\begin{figure}[t]
\begin{center}
\includegraphics[width=0.95\linewidth]{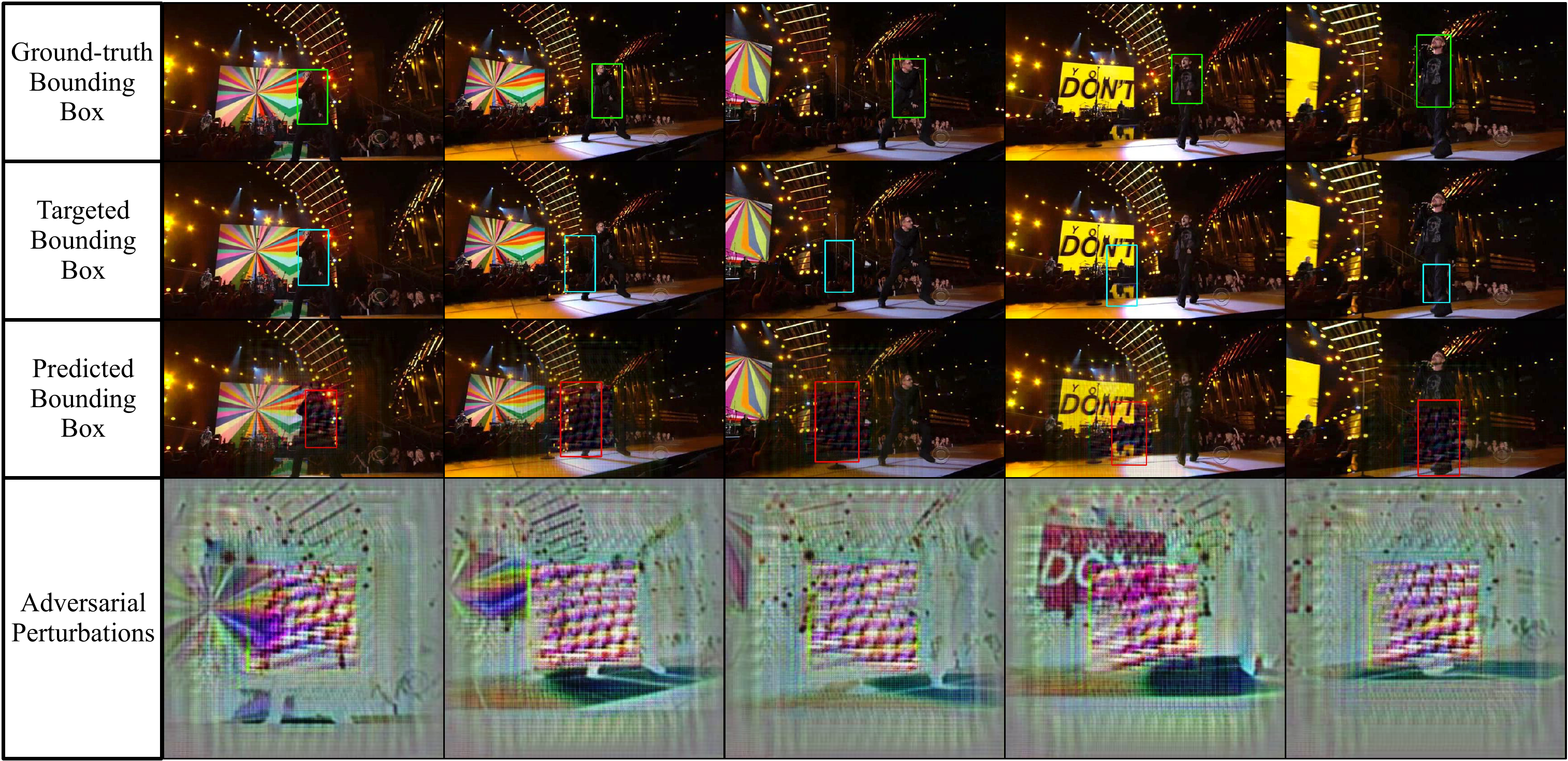}
\end{center}
\caption{The results under targeted attacks. Green represents a ground-truth bounding box, cyan represents the specific bounding box, and red represents the predicted bounding box by trackers. The cyan and red boxes are basically the same in time series, which indicates that targeted attack is successful.}
\label{fig44}
\end{figure} 
\\ \indent We visualize the results of the targeted attack in Figure~\ref{fig44}. The first and third lines represent bounding boxes on the clean video and the adversarial video. The second line represents the specific trajectories we automatically generated according to Eq~\ref{eq10}. It can be seen that the predicted bounding box by tracker is basically the same as the specific bounding box. The fourth line shows adversarial perturbations from the search region, which is significantly stronger than adversarial perturbations in the untargeted attack. Therefore, the targeted attack is more difficult than the untargeted attack under limited disturbance.
\subsection{Transferability to SiamRPN}
We use SiamRPN~\cite{li2018high}, SiamRPN+CIR~\cite{zhang2019deeper}, SiamRPN++~\cite{li2019siamrpn++} as black-box attack models to verify the transferability of adversarial examples generated by FAN. SiamRPN uses an RPN network to perform location regression and classification on the response map. SiamRPN+CIR uses the ResNeXt22 network to replace SiamRPN's Alexnet. SiamRPN++ performs layer-wise and depth-wise aggregations to improve accuracy. 
\begin{table}[]
\begin{center}
\begin{tabular}{|c|c|c|c|c|c|}
\hline
\multicolumn{2}{|c|}{Methods} & SiamFC & SiamRPN & \multicolumn{1}{l|}{SiamRPN+CIR} & \multicolumn{1}{l|}{SiamRPN++} \\ \hline
\multirow{3}{*}{OTB2013} & Success Score & 74\% & 55\% & 46\% & 33\% \\ \cline{2-6} 
 & Precision Score & 76\% & 47\% & 58\% & 35\% \\ \cline{2-6} 
 & Success Rate & 81\% & 56\% & 47\% & 35\% \\ \hline
\multirow{3}{*}{OTB2015} & Success Score & 72\% & 44\% & 45\% & 32\% \\ \cline{2-6} 
 & Precision Score & 75\% & 51\% & 58\% & 37\% \\ \cline{2-6} 
 & Success Rate & 81\% & 55\% & 43\% & 39\% \\ \hline
\end{tabular}	
\end{center}
\caption{Transferability of adversarial examples on two datasets.}
\label{tab3}
\end{table}  
\\ \indent The experimental results are shown in Table~\ref{tab3}. The first column refers to the drop rate of a white-box attack method. The other columns refer to the drop rate of black-box attack methods. We find that black-box attack methods have a lower drop rate than the white-box attack method. It is obvious that black-box attack methods are more difficult than a white-box attack method. Except for the precision score, the performance of the black-box attack in SiamRPN is better than SiamRPN+CIR. This may be due to SiamRPN and SiamFC using the same feature extraction network AlexNet. The black-box attack in SiamRPN++ performs the worst. This is because the architecture of SiamRPN++ can correct some spatial offsets. Even in this case, the drop rate of the black-box attacks can still reach 32\%. The results show that our method can still show good transferability for different tracking methods. 
\section{Conclusion}
In this paper, we accomplished the adversarial attacks for the Visual Object Tracking (VOT) task. We analyzed the weaknesses of DNNs based VOT models: the feature networks and the loss function, and then designed different attacking strategies. We firstly presented a drift loss to make the high-score area obtained by adversarial examples be offset with the original area. Then a pre-defined trajectory was embedded into the feature space of the original images to perform the targeted attack. Finally, we proposed an end-to-end framework to integrate these two modules. Experiments conducted on two public datasets verified the effectiveness of the proposed method. In addition, our method not only achieved excellent performance on the white-box attack, but also on the black-box attack, which expanded its application area. Furthermore, the image quality assessment showed that the generated adversarial examples had good imperceptibility, which guaranteed the security of the adversarial examples.
\section*{Acknowledgement} Supported by the National Key R\&D Program of China (Grant No. 2018AAA010\\0600), National Natural Science Foundation of China (No. U1636214, 61861166002, No. 61806109), Beijing Natural Science Foundation (No. L182057), Zhejiang Lab (NO.2019NB0AB01), Peng Cheng Laboratory Project of Guangdong Province PCL2018KP004.
\par\vfill\par

\clearpage
%
%
\bibliographystyle{splncs04}
\bibliography{egbib.bib}
\end{document}